\documentclass[11pt]{article}
\usepackage[hyperref]{ccl2022-en}
\usepackage{times}
\usepackage{url}
\usepackage{latexsym}
\usepackage{fancyhdr}
\usepackage{graphicx}
\usepackage{amsmath,amssymb,amsfonts}
\usepackage{algorithmic}

\title{Low-Resource Named Entity Recognition Based on Multi-hop Dependency Trigger}

\author{Jiangxu Wu* \\
 {\tt wujx27@mail2sysu.edu.cn} \\\And
 Peiqi Yan* \\
 {\tt yanpeiqiswu@163.com} \\}

\date{}

\begin{document}
\maketitle
\begin{abstract}
This paper introduces DepTrigger, a simple and effective model in low-resource named entity recognition (NER) based on multi-hop dependency triggers. Dependency triggers refer to salient nodes relative to an entity in the dependency graph of a context sentence. Our main observation is that triggers generally play an important role in recognizing the location and the type of entity in a sentence. Instead of exploiting the manual labeling of triggers, we use the syntactic parser to annotate triggers automatically. We train DepTrigger using an independent model architectures which are Match Network encoder and Entity Recognition Network encoder. Compared to the previous model TriggerNER, DepTrigger outperforms for long sentences, while still maintain good performance for short sentences as usual. Our framework is significantly more cost-effective in real business.
\end{abstract}

\section{Introduction}
\label{intro}

%
%
\cclfootnote{
    %
    %
    \hspace{-0.65cm}  
     \noindent * Equal contribution 
    
    \noindent \textcopyright 2022 China National Conference on Computational Linguistics
    
    \noindent Published under Creative Commons Attribution 4.0 International License
}
Named Entity Recognition (NER) aims to detect the span from text belonging to the semantic category such as person, location, organization, etc. NER plays a core component in many NLP tasks and is widely employed in downstream applications, such as knowledge graph ~\cite{Ji:2021}, question answering ~\cite{Molla:2004}  and dialogue system ~\cite{Peng}. The deep-learning based approaches have shown remarkable success in NER, while it requires large corpora annotated with named entities. Moreover, in many practical settings, we wish to apply NER to domains with a very limited amount of labeled data since annotating data is a labor-intensive and time-consuming task. Therefore, it is an emergency to improve the performance of the deep-learning based NER model with limited labeled data.

Previous work in low-resource NER mainly focused on meta-learning ~\cite{Snell}, distantly supervision ~\cite{Yang:2018}, transfer learning ~\cite{Lin:2017},et al. Recently, ~\cite{CLin:2020} proposed an approach based on entity trigger called \emph{TriggerNER}. The key idea is that an entity trigger is a group of words that can help explain the recognition process of an entity in a sentence. Considering the sentence “Biden is the president of \textunderscore”, we are able to infer that there is a country entity on the underline according to “the president of”. In this case, “the president of” is a group of triggers. Experiments reveal that the performance of utilizing 20\% of the trigger-annotated sentences is comparable to that of exploiting 70\% of conventional annotated sentences. However, crowd-sourced entity trigger annotations, which suffer from the same problem as traditional annotation, require labor costs and expert experience.

\begin{figure}[htbp]
\centerline{\includegraphics[width=0.7\textwidth]{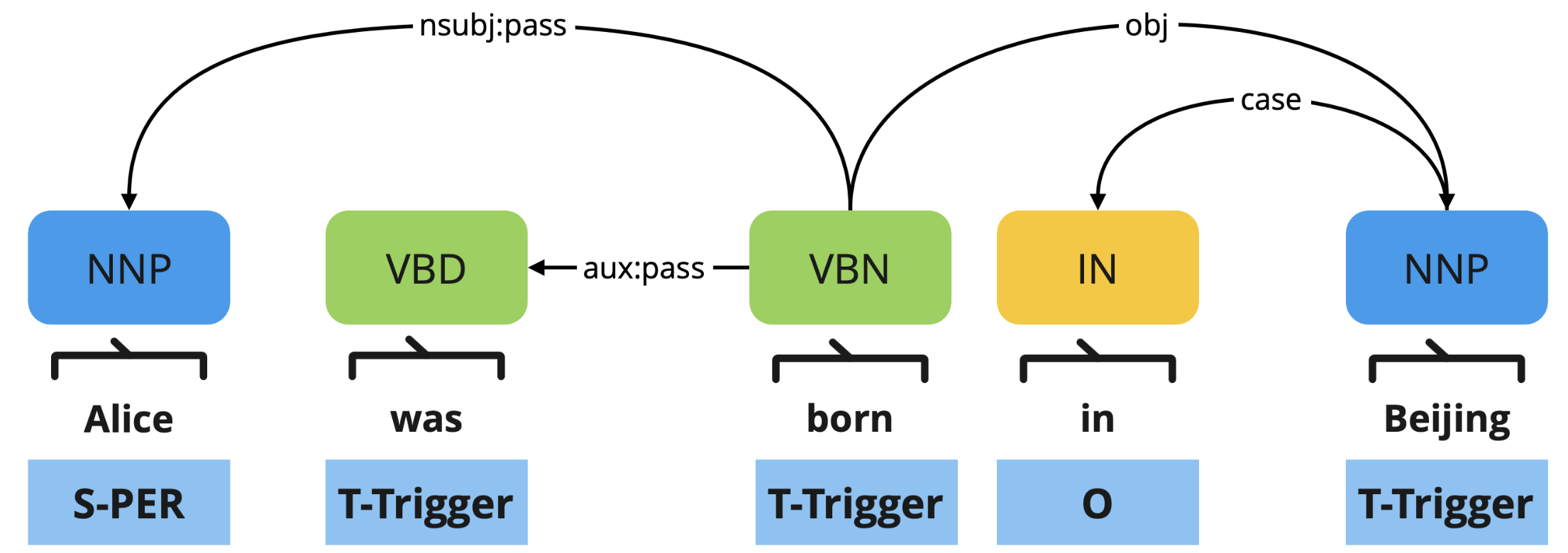}}
\caption{The dependency parse results of "Alice was born in Beijing", "S-PER" is entity label, "T-Trigger" is trigger label, "O" denotes others.}
\label{Fig 1}
\end{figure}

Inspired by attribute triggers in Attribute Extraction ~\cite{Huang2:2017}, this paper presents an alternative approach to automatically annotate the trigger in a sentence by utilizing the syntactic parse. Fig.~\ref{Fig 1}  is the dependency parse result of the sentence “Alice was born in Beijing”, the relation “nsubj:pass” shows that the subject of “born” is “Alice”. According to the meaning of “born”, we are capable of inferring that “Alice” is a person entity. Inspired by this fact, we propose a novel model, namley \emph{DepTrigger}, which explore the structures of dependency trees and utilize the syntactic parser to annotate trigger in a sentence.

Naturally, we propose a simple yet effective framework for low-resource NER, namely \emph{DepTriggerNER}. It includes a trigger semantic matching module (Trigger Match Network) and a sequence annotation module (Entity Recognition Network). The DepTriggerNER adopts two-steps pipeline mode: 1) we first trains the Trigger Match Network module for learning trigger representation; and 2) we combine trigger representation to train the Entity Recognition Network module. Our main contribution includes the new proposed "DepTrigger" model, which reduces the cost and complexity by using a syntactic parser to automatically annotate trigger. 

We evaluate DepTrigger on CoNLL2003 \cite{Erik:2003} and BC5CDR \cite{Li:2016}, where DepTrigger outperforms the TriggerNER model on BC5CDR but slightly under-performs on CoNLL2003. Compared to TriggerNER, DepTrigger is particularly useful in its ability to automatically produce annotated triggers. Besides, the independent model architectures have a better performance. Our results suggest that DepTrigger is a promising alternative to the TriggerNER in low-resource NER tasks.

\begin{figure}[htbp]
\centering
\includegraphics[width=0.6\textwidth]{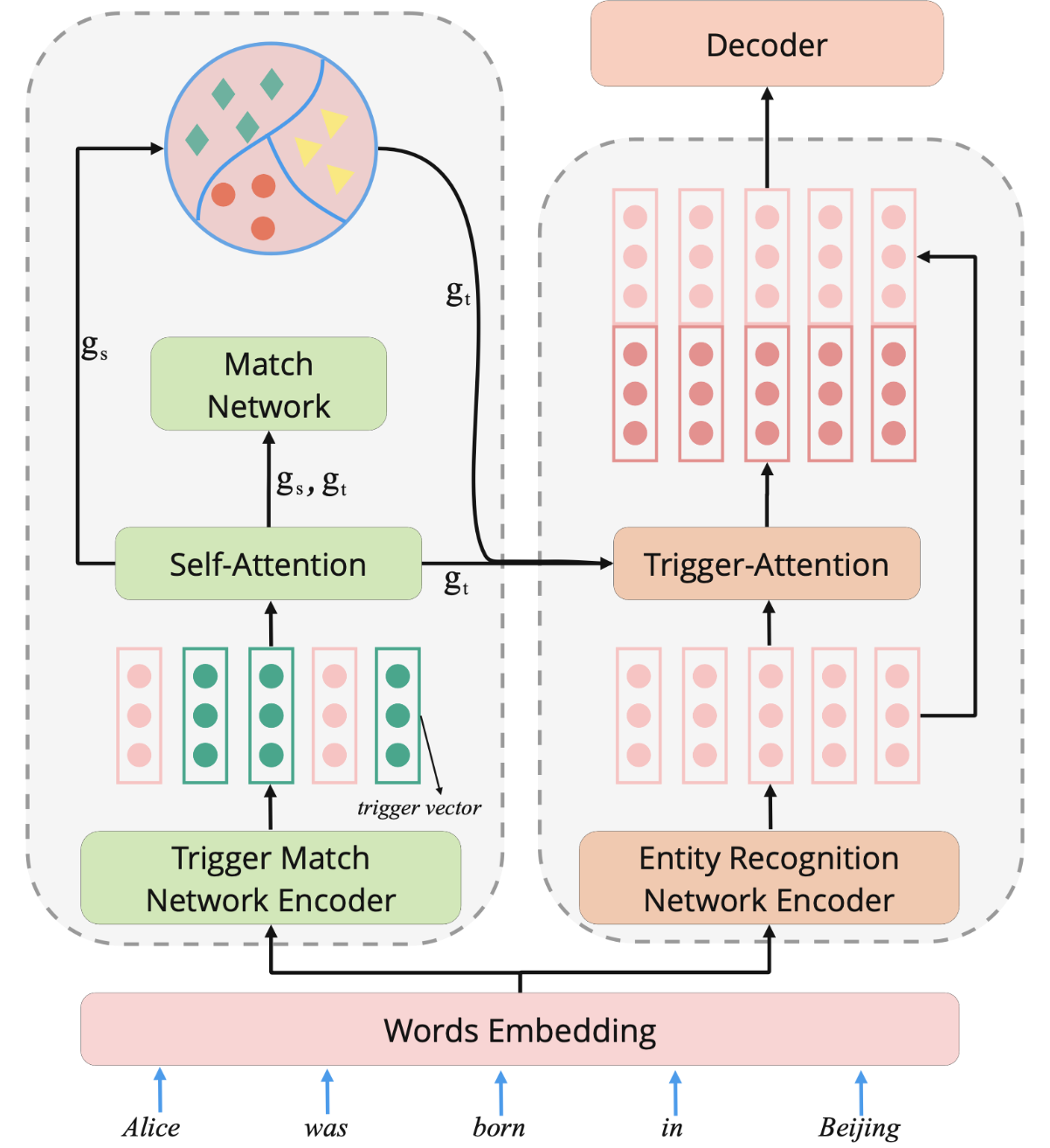}

\caption{The framework of DepTriggerNER. The left is the Trigger Match Network. The right is the Entity Recognition Network. The circle in the upper left corner is Trigger Pattern Prototype, it is a look-up table generated by Trigger Match Network after training.}
\label{Fig 2}
\end{figure}

\section{Model}
In this section, we present the framework of DepTriggerNER in Fig.~\ref{Fig 2}. Compared to TriggerNER, there are three main differences: (1) instead of crowd-sourced, we harness the syntactic parser to annotate trigger automatically; (2) we omit the trigger classification network; and (3) the Trigger Match Network encoder and the Entity Recognition Network encoder are independent of each other.

This section is organized as follows. We first describe how to use the syntactic parser to annotate dependency trigger in section 2.1. We then introduce Trigger Match Network and Entity Recognition Network in section 2.2 and section 2.3, respectively.

\subsection{DepTrigger}
DepTrigger are prominent nodes relative to an entity in the context sentence dependency graph. We apply Stanford CoreNLP to the sentences to obtain dependency paths. The dependency paths is a directed graph with words as nodes and dependencies as edges. Fig. ~\ref{Fig 1} shows the dependency parse results of the sentence “Alice was born in Beijing”.  In Fig.~\ref{Fig 1} , “born” is connected with the entity “Alice” by relation “nsubj:pass”, so that “born” is a DepTrigger. Words have a one-hop relationship with entities are called primary triggers, and words have a two-hop relationship with entities are called secondary triggers.

\subsection{Trigger Match Network}
Each entity contains a group of DepTrigger, which form a trigger pattern. We assume that each sentence has an entity and contains a trigger pattern. In the training stage, the Trigger Match Network aims to learn the representation of trigger patterns and sentences. In the inference stage, the trigger pattern representation with similar semantics to the sentence representation will be selected from the Trigger Pattern Prototype.

In Fig.~\ref{Fig 2}, each sentence is first transformed into a vector by the Words Embedding module. Then, the hidden state matrix is obtained through the Trigger Match Network Encoder. The self-attention layer is used to obtain sentence representation $\vec{g}_s$  and trigger pattern representation $\vec{g}_t$,  ~\cite{Lin:2017} defined as follows:
\begin{gather}%
	\vec{\alpha}_s=Softmax\left( W_2 \times tanh\left( W_1 \times H \right) \right)\\
    \vec{g}_s=\vec{\alpha}_s H\\
    \vec{\alpha}_t=Softmax\left(W_2 \times tanh\left(W_1 \times M\right)\right)\\
    \vec{g}_t=\vec{\alpha}_t M
\end{gather}

$W_1$ and $W_2$ are the trainable parameters. $H$ and $M$  represent the hidden state matrix of the sentence and the hidden state matrix of DepTrigger, respectively.

The Match Network calculates the distance between trigger pattern representation and sentence representation. The matching loss function ~\cite{CLin:2020}  is defined as follows:

\begin{equation}
	L = \begin{cases}
	||\vec{g}_s-\vec{g}_t||_2^2, t \in s\\
	max\left(0,m-||\vec{g}_s-\vec{g}_t||_2^2\right),t \notin s
		   \end{cases}
\end{equation}

$||\cdot||_2$  is L2-norm distances, $m$ is margin. $t \in s$ indicates trigger pattern representation and sentence representation matches well while $t \notin s$ is on the contrary. We create negative samples by randomly matching trigger pattern representation and sentence representation in a batch.

\subsection{Entity Recognition Network}
Entity Recognition Network is similar to most deep-learning based NER models and consists of encoder and decoder. However, the Entity Recognition Network has been added a trigger-attention layer. Note that the parameters of Trigger Match Network are frozen when training Entity Recognition Network.

In training, each sentence passes through the Trigger Match Network Encoder and the Entity Recognition Network Encoder, respectively. Then, $\vec{g}_t$ is obtained from the self-attention layer. In the trigger-attention layer, $\vec{g}_t$ is used to calculate the weight of each vector in the Entity Recognition Network Encoder’s outputs as follows ~\cite{Luong:2015}:
\begin{gather}
	\vec{\alpha}=Softmax\left( \vec{v}\times tanh\left( U_1 \times H + U_2 \times \vec{g}_t \right) \right)\\
    H'=\vec{\alpha} H
\end{gather}
$U_1,U_2,\vec{v}$  are model parameters, and $H$  is the Entity Recognition Network Encoder’s outputs matrix. Finally, we concatenate the matrix $H$  with the trigger-enhanced matrix $H$  as the input $\left(\left[H; H'\right]\right)$ fed into the decoder.

\subsection{Inference}
After training, each sentence in the training set is re-input into Trigger Match Network to obtain trigger pattern representation.  We then save these representations in memory, shows as the Trigger Pattern Prototype in Fig.~\ref{Fig 2}. In the inference stage, We first obtain sentence representations $\vec{g}_s$ through Trigger Match Network and then retrieve the semantic similarity vector $\vec{g}_t$ from Trigger Pattern Prototype. Vector $\vec{g}_t$ is used as the attention query in Entity Recognition Network.

\section{Experiments}

\subsection{Experiments Setup}

\begin{table}[htbp]
\centering
\begin{tabular}{c c c c}
\hline
\textbf{Dataset} & \textbf{\#Class} & \textbf{\#Sent} & \textbf{\#Entity}\\
\hline
{CoNLL'03} & {4} & {14986} & {23499} \\
{BC5CDR} & {2} & {4560} & {9385} \\\hline
\end{tabular}
\caption{Data statistics.}
\label{Table 1}
\end{table}

CoNLL2003 \cite{Erik:2003} and BC5CDR \cite{Li:2016} are used to evaluate our model. The statistics of these datasets are shown in Table.~\ref{Table 1}.
We choose BiLSTM-CRF ~\cite{Ma:2016}  and TriggerNER ~\cite{CLin:2020} as baseline models. TriggerNER is the first trigger-based NER model. We choose BiLSTM as encoder and CRF as decoder in our model. To ensure a fair comparison, we use the same codebase and words embedding from GloVE  ~\cite{Pennington:2014}, which used in baseline model. The hyper-parameters of the model are also the same. Our code and data are released \footnote{https://github.com/wjx-git/DepTriggerNER}.

We choose BIOES tagging schema for non-triggers, and triggers are all labeled with ``T-trigger''. In order to make the model learn the relation between entity and its trigger better, we repeat a sentence N times, and N is the number of entities in the sentence. Each sentence retains one entity and its trigger, other entities are marked as non-entities. 

\begin{table*}[htbp]
\centering
\begin{tabular}{ p{1cm} p{1.5cm}|p{1cm} p{1.2cm} p{1cm} | p{1cm} p{1.5cm}|p{1cm} p{1.2cm} p{1cm} }
\hline
\multicolumn{5}{c|}{CoNLL 2003} & \multicolumn{5}{c}{BC5CDR} \\
\hline
\#sent &BiLSTM-CRF &\#trig &Trigger-NER &Ours &\#sent &BiLSTM-CRF &\#trig &Trigger-NER &Ours\\
\hline
 5\%  &69.04    &3\%  &75.33 &\textbf{77.42} &5\% &71.87	&3\% &61.44 &\textbf{63.37}\\
 10\% &76.83	&5\% &80.2  &\textbf{80.26} &10\% &72.71	&5\% &66.11 &\textbf{66.92}\\
 20\% &81.3 	&7\%  &\textbf{82.02}  &81.3 &20\% &69.92	&7\% &67.22 &\textbf{69.27}\\
 30\% &83.23	&10\%   &\textbf{83.53} &82.96 &30\% &73.71	&10\% &70.71 &\textbf{71.42}\\
 40\% &84.18	&13\%   &\textbf{84.22}  &83.26 &40\% &72.71	&13\% &71.87 &\textbf{73.17}\\
 50\% &84.27	&15\%   &\textbf{85.03} &83.86 &50\% &75.84	&15\% &71.89 &\textbf{74.35}\\
 60\% &85.24    &17\%  &\textbf{85.36} &84.32 &60\% &75.84	&17\% &73.05 &\textbf{75.08}\\
 70\% &86.08    &20\%  &\textbf{86.01} &84.53 &70\% &76.12	&20\% &73.97 &\textbf{76.44}\\
\hline
\end{tabular}
\caption{F1 score results. ``\#sent'' denotes the percentage of the sentences labeled only with entity label, ``\#trig'' denotes the percentage of the sentences labeled with both entity label and trigger label.}
\label{Table 2}
\end{table*}

\subsection{Results}
\begin{figure}[htbp]
\centering
\includegraphics[width=0.6\linewidth]{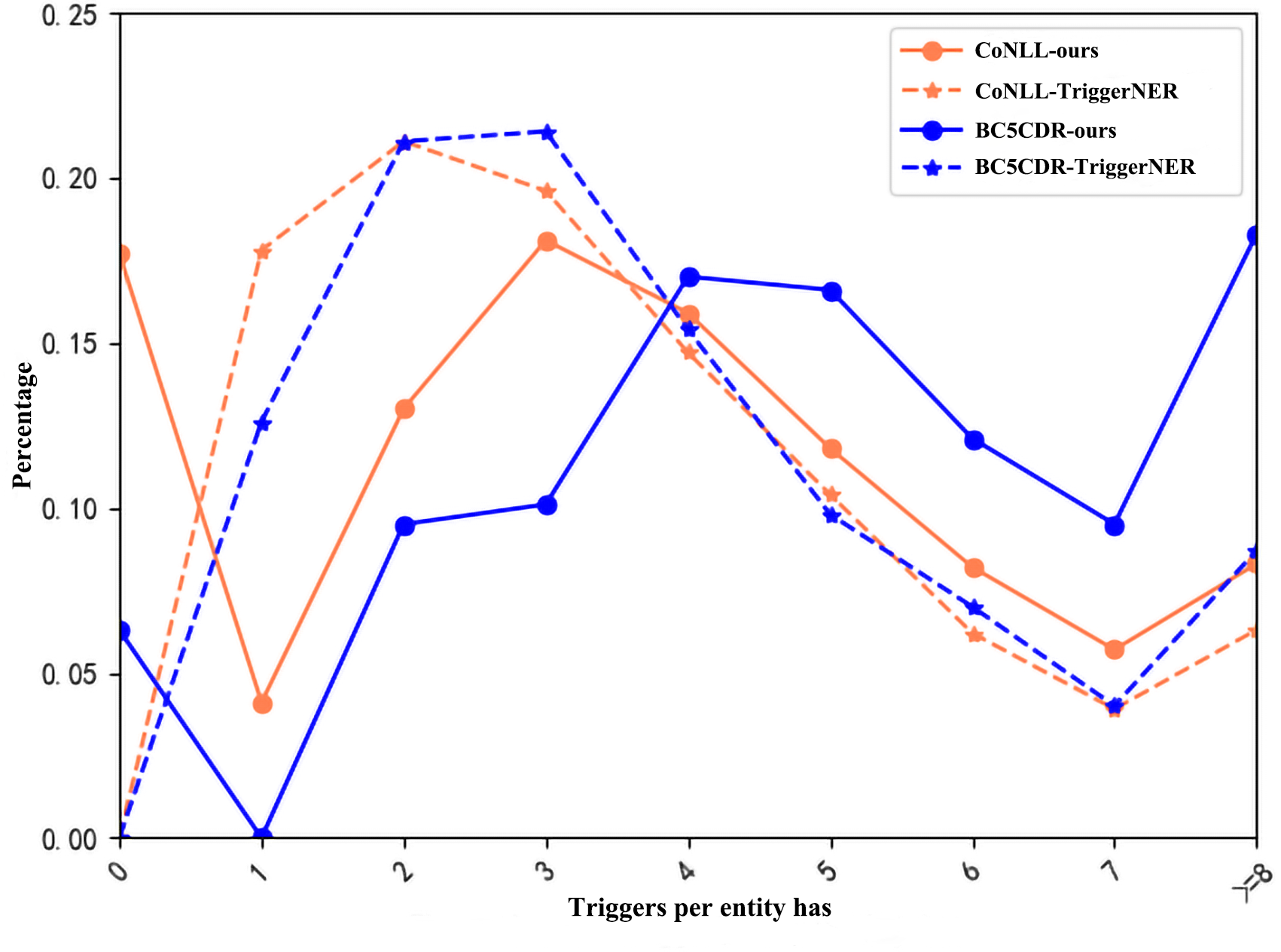}
\caption{Ratio of the number of sentences with the number of triggers each entity has in the dataset. The X-axis is the number of triggers of a entity has, and the Y-axis is the percentage. The solid lines represent the trigger of ours. The yellow line represents CoNLL datasets.}
\label{Fig 3}
\end{figure}

As shown in Tabel.~\ref{Table 2},  Our model achieves a similar performance as TriggerNER. More detailed, our model performs better on BC5CDR than TriggerNER, but slightly worse on CoNLL2003. We explain this phenomenon in terms of the number of triggers each entity has. Fig.~\ref{Fig 3} shows the ratio of the number of sentences with the number of triggers an entity has in each dataset. The two yellow curves are very close when the abscissa value is greater than 3, and the yellow dotted line is larger than the solid line when the abscissa value is less than 3. This fact demonstrates that on CoNLL2003 the number of triggers annotated by our method is less than TriggerNER. In the two blue curves, the solid blue line is larger than the dashed line when the abscissa value is greater than 4, and the opposite is true when the abscissa value is less than 4. This shows that the number of triggers annotated by our method is more than TriggerNER on BC5CDR. We believe that an entity is easier to recognize when it has more triggers, which would explain why our model performs better on BC5CDR and slightly worse on CoNLL2003.

We analyzed the sentence length distribution in the two datasets to further understand why we annotate fewer triggers in CoNLL and more in BC5CDR than in TriggerNER. The statistical results of sentence length distribution in Table ~\ref{Table 3}, show that sentences are shorter in the CoNLL dataset and longer in the BC5CDR dataset. From Table ~\ref{Table 3} and Figure ~\ref{Fig 3}, it can be concluded that our method can label more triggers in long sentences but fewer triggers in short sentences compared to manual marking in TriggerNER. Therefore, our method is more suitable for datasets with longer sentences.

\begin{table}[htbp]
\centering
\begin{tabular}{ p{1.3cm}| p{1cm}| p{1cm} |p{1cm}| p{1cm} }
\hline
 Datasets &1\textasciitilde10 &10\textasciitilde25 &25\textasciitilde50  &50\textasciitilde \\
\hline
 CoNLL  &52.32\% &27.33\%  &19.93\%   &0.42\% \\
 \hline
 BC5CDR &5.7\%	&50.64\%   &37.54\%  &6.51\%\\
\hline
\end{tabular}
\caption{Statistical results of sentence length distribution}
\label{Table 3}
\end{table}

\begin{table}[htbp]
\centering
\begin{tabular}{ p{1cm}| p{1cm} p{1.2cm}| p{1cm} p{1.2cm} }
\hline
{\#trig} &
\multicolumn{2}{c|}{CoNLL 2003} & \multicolumn{2}{c}{BC5CDR} \\
\cline{2-5}
 &merge &separate &merge  &separate \\
\hline
 3\%  &76.36    &\textbf{77.42}   &61.3   &\textbf{63.37}\\
 5\% &79.38	    &\textbf{80.26}   &66.15  &\textbf{66.92}\\
 7\% &80.37 	&\textbf{81.3}    &68.02  &\textbf{69.27}\\
 10\% &81.58	&\textbf{82.96}   &70.93  &\textbf{71.42}\\
 13\% &82.55	&\textbf{83.26}   &72.7   &\textbf{73.17}\\
 15\% &83.03	&\textbf{83.86}   &73.25  &\textbf{74.35}\\
 17\% &83.51    &\textbf{84.32}   &74.95  &\textbf{75.08}\\
 20\% &83.81    &\textbf{84.53}   &75.08  &\textbf{76.44}\\
\hline
\end{tabular}
\caption{Comparative experiment F1 score results. \textbf{merge} means to merge Trigger Match Network encoder and Entity Recognition Network encoder. \textbf{separate} means to separate Trigger Match Network encoder and Entity Recognition Network encoder. The best results are in \textbf{bold}.}
\label{Table 4}
\end{table}
In our model, Trigger Match Network encoder and Entity Recognition Network encoder are independent, which is different from the TriggerNer. The main purpose of Trigger Match Network is to learn the representation of trigger patterns, and Entity Recognition Network is to learn entity representation. So we think we can not get an advantage by combining Trigger Match Network and Entity Recognition Network because they need to capture specific information. That is inspired by ~\cite{Zexuan:2021}, and they observe that the contextual representations for the entity and relation models essentially capture specific information, so sharing their representations hurts performance.

We do a comparative experiment to test the performance of our model for merging and separating, respectively, while leaving everything else unchanged. The experimental results are shown in Table.~\ref{Table 4}, \textbf{merge} means to merge Trigger Match Network encoder and Entity Recognition Network encoder. Separate means to separate Trigger Match Network encoder and Entity Recognition Network encoder. It shows that the performance is better when the Trigger Match Network encoder and Entity Recognition Network encoder are independent.

In order to compare the influence of primary and secondary trigger words on the model, we backup two datasets of CoNLL, and only the primary triggers are labeled in one dataset, and only the secondary trigger words are labeled in the other dataset, do the same for BC5CDR. Table.~\ref{Table 5}  shows the F1 score on these datasets. Compared primary and secondary trigger, there is no evident show that one is better than the other. 
Combined with table ~\ref{Table 1} and table ~\ref{Table 4}, the effect of using the primary trigger and the secondary trigger at the same time is significantly better than that of using them alone.
.
\begin{table}[htbp]
\centering
\begin{tabular}{ p{0.8cm}| p{1cm}  p{1.4cm}| p{1cm} p{1.4cm} }
\hline
{\#trig} &
\multicolumn{2}{c|}{CoNLL 2003} & \multicolumn{2}{c}{BC5CDR} \\
\cline{2-5}
 &primary &secondary &primary &secondary \\
\hline
 3\%  &\textbf{63.4}     &62.35   &\textbf{52.3}   &50.92\\
 5\% &\textbf{66.3}	    &\textbf{66.3}   &54.17  &\textbf{55.84}\\
 7\% &\textbf{70.37} 	&69.44    &\textbf{58.92}  &57.33\\
 10\% &\textbf{74.02}	&73.44   &\textbf{60.32}  &60.24\\
 13\% &74.86        	&\textbf{74.91}   &61.35   &\textbf{62.01}\\
 15\% &\textbf{76.2}	&75.46   &64.26  &\textbf{64.25}\\
 17\% &\textbf{77.36}    &76.33   &\textbf{64.51}  &64.26\\
 20\% &\textbf{77.55}    &77.53   &65.94  &\textbf{66.69}\\
\hline
\end{tabular}
\caption{Comparative experiment of primary and secondary trigger}
\label{Table 5}
\end{table}

\section{Conclusion and Future Work}

We have introduced dependency trigger to incorporate trigger information into NER method. The core of our method is using syntactic parser to automatically label the trigger of entities. Our model performs well for long sentences, while maintain similar performance as TriggerNER for short sentences. Thanks to automatically annotate trigger of entities, our framework is more practical in the real business. 	Future work with DepTrigger includes: 1) adjusting our model to encoder based on language model; 2) making a further analysis of trigger type; 3) developing models for improving the performance on short sentences.


\end{document}